\documentclass[letterpaper, 10 pt, conference]{ieeeconf}  
\usepackage{amsmath,amssymb}
\usepackage{graphicx}
\usepackage{bbm} 
\usepackage{booktabs}
\usepackage{xcolor} 
\usepackage{hyperref}
\usepackage{cleveref}
\usepackage{cite}
\usepackage{dsfont}
\usepackage{color, colortbl}
\definecolor{gray}{rgb}{0.95, 0.95, 0.96}
\graphicspath{ {./fig/} }
\makeatletter
    \def\tagform@#1{\maketag@@@{\normalsize(#1)\@@italiccorr}}
\makeatother

\IEEEoverridecommandlockouts                              

\newcommand{\best}[1]{{\textbf{{#1}}}}
\newcommand{\first}[1]{{\color{blue}{#1}}}
\newcommand{\second}[1]{{\color{orange}{#1}}}
\newcommand{\third}[1]{{#1}}

\newcommand{\mypar}[1]{\noindent\textbf{#1}}

\title{\LARGE \bf
Narrowing the coordinate-frame gap in behavior prediction models: \\Distillation for efficient and accurate scene-centric motion forecasting
}

\author{DiJia (Andy) Su$^{1}$, Bertrand Douillard$^{2}$, Rami Al-Rfou$^{2}$, Cheolho Park$^{2}$,  Benjamin Sapp$^{2}$
\thanks{$^{1}$Princeton University, $^{2}$Waymo LLC}%
}

\begin{document}

\maketitle
\thispagestyle{empty}
\pagestyle{empty}

\begin{abstract}

Behavior prediction models have proliferated in recent years, especially in the popular real-world robotics application of autonomous driving, where representing the distribution over possible futures of moving agents is essential for safe and comfortable motion planning. In these models, the choice of coordinate frames to represent inputs and outputs has crucial trade offs which broadly fall into one of two categories. {\em Agent-centric} models transform inputs and perform inference in agent-centric coordinates. These models are intrinsically invariant to translation and rotation between scene elements, are best-performing on public leaderboards, but scale quadratically with the number of agents and scene elements. {\em Scene-centric} models use a fixed coordinate system to process all agents. This gives them the advantage of sharing representations among all agents, offering efficient amortized inference computation which scales linearly with the number of agents. However, these models have to learn invariance to translation and rotation between scene elements, and typically underperform agent-centric models.

In this work, we develop knowledge distillation techniques between probabilistic motion forecasting models, and apply these techniques to close the gap in performance between agent-centric and scene-centric models. This improves scene-centric model performance by 13.2\% on the public Argoverse benchmark, 7.8\% on Waymo Open Dataset and up to 9.4\% on a large In-House dataset. These improved scene-centric models rank highly in public leaderboards and are up to 15 times more efficient than their agent-centric teacher counterparts in busy scenes.

\end{abstract}


\section{INTRODUCTION AND RELATED WORK}
\begin{figure}[t!]
\centering
\includegraphics[width=1.0\columnwidth]{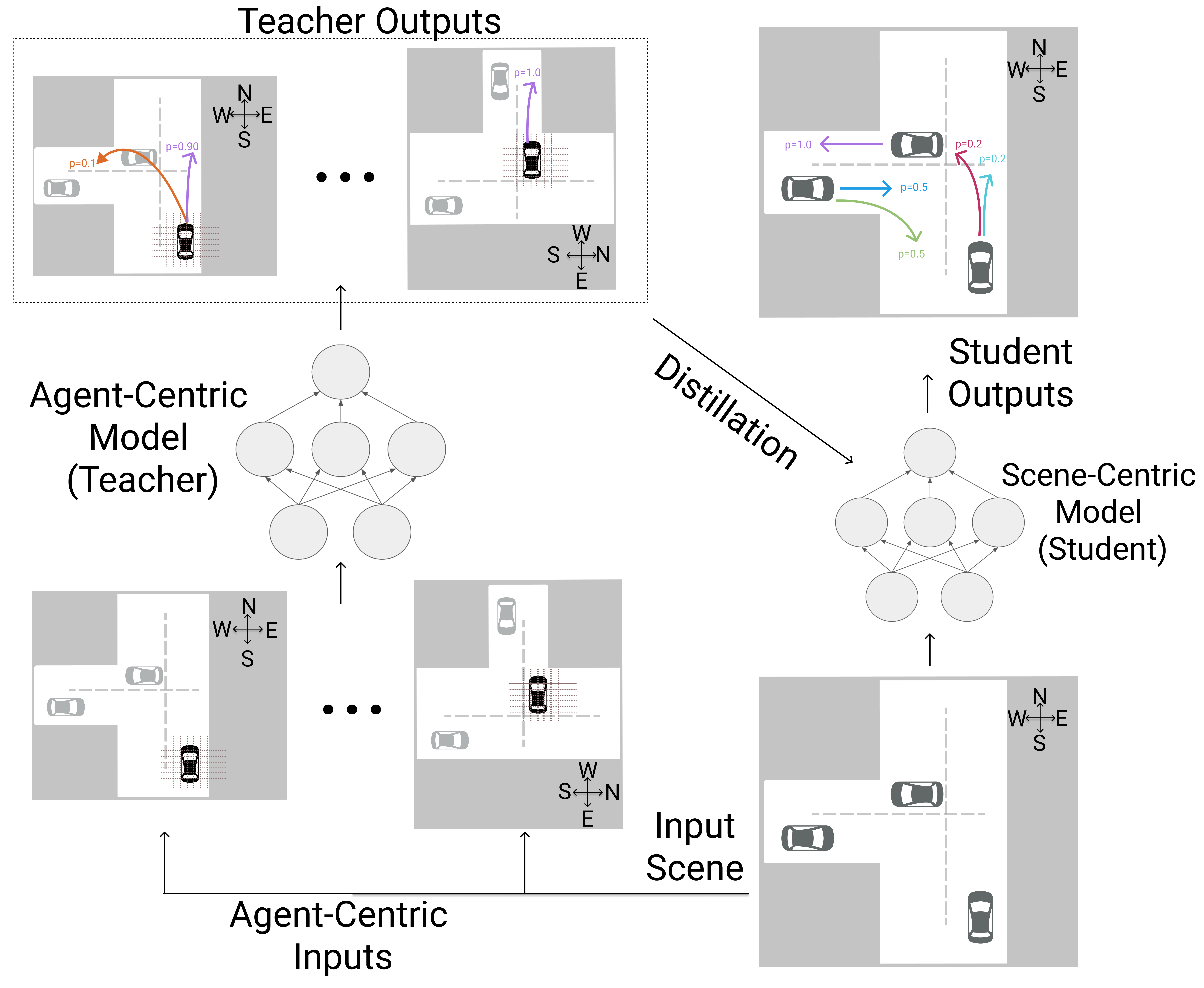}
\caption{Approach overview. On the left, the teacher and agent-centric model is repeatedly and independently applied to each agent in the scene, with all model inputs and outputs represented in each agent's ego-centric coordinate frame. On the right, the student and scene-centric model is applied to the whole scene once, without requiring repeated computations per agent. While faster, a scene-centric formulation tend to be less accurate, since it also has to understand and model the per-agent invariance that is otherwise built-in the agent-centric approach. To use the computational efficiency of a scene-centric approach and yet benefit from the accuracy of an agent-centric approach we propose a knowledge distillation approach that uses the predicted trajectories of the agent-centric or teacher model, to train a scene-centric or student model.\vspace{-15pt}}
\label{fig:system_overview}
\end{figure}

Predicting the future behavior of multiple vehicle, cyclist, and pedestrian agents in real-world driving scenes is a difficult but essential task for safe and comfortable motion planning for autonomous vehicles. This task is typically referred to as ``motion forecasting" or ``behavior prediction". It is challenging for a number of reasons. (1) The world state is heterogeneous, consisting of static and dynamic road network elements and dynamic agent state observations. (2) The outcomes depend heavily on multi-agent interactions. (3) The output distribution over possible futures is inherently uncertain and highly multi-modal due to latent agent intents. How to represent the input world state, interactions, and output distributions are all open questions and active areas of research.

In the last few years, there has been a proliferation of behavior prediction systems which address these modeling challenges, fueled by both the compelling promise of the autonomous vehicle industry, and public benchmarks to compare methods \cite{womd,argoverse, d3, d4}. One of the most interesting design choices and the focus of this paper is that of the coordinate frames to represent input and output data. There are two distinct common choices, each with advantages and disadvantages.

\textbf{Agent-centric} models represent inputs and internal state in agent-centric coordinates, and perform inference reasoning in this frame\footnote[3]{Without loss of generality, an agent-centric frame transforms world coordinates so that the origin is set to the ego-agent's center, and rotated so that the agent's heading direction is the unit vector $(x,y) = (1,0)$.}. The coordinates of road elements (e.g., lanes, crosswalks) and other agents' states are described relative to the agent's pose, thus the representation is inherently invariant to the global position and orientation of the agent. This can be considered a form of feature pre-processing that allows for models to specialize to an agent's point of view, and in practice results in state-of-the-art performance on public benchmarks \cite{a0, a1, a2, a3, a4}. A key downside, however, becomes apparent when modeling many agents in a scene: each agent is modeled independently, thus computation is typically linear in the number of agents, and quadratic when modeling interactions\cite{mercat2020multi,zhao2020tnt,wimp2020,tang_multifuture,casas2020spagnn,precog_Rhinehart_2019_ICCV,salzmann2020trajectron++}---for a scene with $n$ agents, and $m$ road elements, the computation scales as $O(n(n+m))$. This is not an issue for public benchmarks which require modeling of less than ten agents at once \cite{womd,argoverse, d3, d4}, but is a computational bottleneck for busy real-world urban environments consisting of hundreds of agents.

\textbf{Scene-centric} models, on the other hand, do the bulk of world state encoding in a shared, fixed frame for all agents\footnote[4]{Without loss of generality, this can be an arbitrary conceptual center of the scene elements.}. Models that operate in this frame are typically  ``top-down" or ``bird's-eye-view" representations which discretize the world into spatial grid cells, and apply a convolutional neural net (CNN) backbone to encode the scene~\cite{lee2017desire, s0, s1, chai2019multipath, s3, yuan2019diverse, buhet2020plop}---although non-raster scene-centric approaches also exist~\cite{ngiam2021scene,liang2020laneGCN,zeng2021lanercnn}. After such processing, the prediction head of these models decodes trajectories in agent-frame after a global-to-local transformation. A salient advantage of this formulation is that computation is primarily function of the spatial grid resolution and field of view, rather than the number of agents---a spatial grid of size $H \times W$ cells would scale as $O(HW + n)$, where the first term is processed with a CNN and dominates the second term in practical settings---see fig.~\ref{fig:inference} for quantification. The downsides of this formulation are (1) loss of information when discretizing world state into a raster format, (2) difficulty in modeling long-range interactions with CNNs, and (3) the model must either learn rotation/translation invariance or learn to perform a global to local transformation for each agent when decoding.

In brief, agent-centric models outperform scene-centric models---borne out in public leaderboards, and likely explainable by the shortcomings described above. However, scene-centric models are compelling due to amortized sub-linear scaling with respect to the number of agents in the scene; particularly relevant in dense urban environments. In this paper, we propose a novel {\em knowledge distillation} method to narrow the gap in performance between these two different modeling approaches.

{\bf Knowledge distillation} \cite{hinton2015distilling} is a popular and effective machine learning technique in domains like computer vision and natural language processing to transfer knowledge from a large model---the ``teacher model"---to a smaller one--the ``student model".  The knowledge transfer mechanism originally proposed for classification tasks, replaces training data groundtruth (``hard labels") with predictions from the teacher model (``soft labels"). The intuition is that these soft labels contain a more information-rich smooth target space for the student model to learn from than the original data~\cite{phuong2021understanding}\cite{cheng2020explaining}.

Distillation has been extended beyond classification to sequence prediction tasks like Neural Machine Translation~\cite{Kim2016SequenceLevelKD}. To our knowledge, however, distillation has never been applied to the domain of behavior prediction / motion forecasting. 
Although behavior prediction can be considered a sequence problem, a key difference is that we wish the predicted future distributions to cover the entire space of outcomes accurately, in contrast to the typical NLP task for instance that aims to generate a single realistic output. Hence transferring knowledge between a teacher and a student for motion forecasting is an open problem.
Furthermore, for motion forecasting where the future is represented as a distribution of trajectories covering intent modes (the approach we adopt here and is the most common, eg.\cite{chai2019multipath, tang2019multiple, zhao2019multiagent, phanminh2020covernet}), trajectory and mode diversity is crucial.
One key challenge then is that distillation could be detrimental to diversity; this was investigated in \cite{Zhou2020UnderstandingKD} in the NLP domain.

In this work, we develop and empirically validate a variety of distillation approaches for behavior prediction. We then apply these techniques by setting the teacher to be a high-performance agent-centric model, and transfer knowledge to an efficient student scene-centric model. In doing so we significantly improve the performance of our scene-centric model while maintaining it's computational efficiency benefits.

{\bf Contributions.} The contributions of this paper are as follows:

\mypar{$\bullet$} We systematically analyze latency and quality of agent-centric and scene-centric model approaches in a common framework. This supports our characterization of the coordinate-frame modeling choices in an empirically rigorous way.

\mypar{$\bullet$} We are the first to develop and apply knowledge distillation techniques to the popular field of behavior prediction modeling.

\mypar{$\bullet$} Applying our best distillation approach gives a remarkable boost in performance to our efficient scene-centric models on several large autonomous vehicle future prediction datasets. Comparing to the non-distilled student model baseline, distillation improves performance by 13.2\% on the Argoverse dataset, 7.8\% on the Waymo-Open-Motion dataset, and up to 9.4\% on key metrics of our In-House dataset.

\section{Background}

\textbf{Definition of the prediction problem}.
Let $x$ be the observations of all agents in the scene (in the form of past trajectories) and additional contextual information (such as lane semantics and traffic light states), $t$ be the discrete time step, $s_t$ be the state of an agent at time $t$. The future trajectory $s = [s_1, ..., s_T]$ is the sequence of states of the agent up to time $T$. We assume our model predicts $K$ trajectories, where each trajectory is a sequence of predicted states ${s}^k = [{s}^k_1, ..., {s}^k_T]$.

For both agent-centric and scene-centric approaches, we consider the class of models whose output is to predict a Gaussian distribution around a predicted trajectory:
\begin{equation}
    \phi(s^k_t| x) = \mathcal{N} (s^k_t | \mu^k_t(x), \Sigma^k_t(x))
    \label{eqn:gauss}
\end{equation}
where $\mu^k_t$ is the mean and $\Sigma^k_t$ is the co-variance of the Normal distribution. The mean and the variance are learnt parameters. The mean represents the mode of the distribution, which is the most likely state at time $t$.

We also model a probabilistic distribution over the predicted trajectories, which can be interpreted as the ``confidence" over each predicted trajectory: $\pi({s}^k | x) = \frac{e^{f_k(x)}}{\sum_i e^{f_i(x)}}$, where $f_k(x) : \mathbb{R}^{d(x)} \rightarrow \mathbb{R}$ is the output parameterized by a neural network.

Thus, combining the two elements above we obtain the Gaussian Mixture Model (GMM) distribution:
\begin{equation}
    p(s|x) = \sum_{k=1}^{K} \pi({s}^k | x)  \prod_{t=1}^{T} \phi(s_t | {s}^k, x) 
\label{eqn:GMM}
\end{equation}

This makes the simplifying assumption that time steps are conditionally independent given a history of world state, allowing us to use an efficient feed-forward neural network.
A typical number for $K$ is on the order of $K=10$ output trajectories. This type of output representation is a fairly popular approach, as in~\cite{chai2019multipath, tang2019multiple, hong2019rules, salzmann2021trajectron}.

\subsection{Teacher model: Agent-Centric Model}
\label{sec:acm}
We use an {\em agent-centric coordinate frame model} (ACM) to serve as the teacher. The agent-centric model encodes, processes, and reasons about the world from each individual agent's point of view. This representation requires a transformation of all scene information from the global coordinate frame into the agent's frame. Because of this, with the agent-centric approach, inference time and memory requirement increases with the number of agents.

Our ACM architecture is inspired by some best performing design choices in the literature. It consumes the following four types of input: road graph and traffic light information, motion history (i.e. agents states history), and agents interactions. For the road graph information, the ACM utilizes polylines to encode the road elements from a 3D high definition map with an MLP (multi-layer perceptron), similar to~\cite{a1,liang2020laneGCN,wimp2020}. For traffic light information, the ACM utilizes a separate LSTM as the encoder. For the motion history, the ACM uses a LSTM to handle a sequence of past observations, and the last iteration of the hidden state is used as history embedding, as in~\cite{mercat2020multi,SocialGAN,sociallstm,salzmann2020trajectron++}, to name a few. For agent interactions, we use a LSTM to encode the neighbors' motion history in an agent centric frame, and aggregate all neighbors' information via max-pooling to arrive to a single interaction. This is a simple form of fully-connected neighbor interaction modeling; other works have explicitly used GNNs~\cite{liang2020laneGCN,casas2020spagnn} and/or attention or max-pooling~\cite{zhao2020tnt,a1,tang_multifuture,salzmann2020trajectron++}. Finally, these four encodings are concatenated together to create an embedding for \emph{each agent} in the agent-centric coordinate frame. This final embedding is converted into a GMM (eqn.\ref{eqn:GMM}) using an MLP based decoder.

\subsection{Student model: scene-centric model}

For the student we use a {\em scene-centric coordinate frame model} (SCM). In our SCM architecture, the input data is represented in a global coordinate frame that is shared across all agents. As mentioned above, one of the benefits of this formulation is that the scene can be processed as a whole, resulting in efficient inference which is invariant to the number of agents.

The SCM consumes three types of inputs.
It takes in road information represented as points augmented with semantic attributes, agents information in the form of points sampled from each agent's oriented box, and traffic light information, also represented as points augmented with semantic attributes.
The SCM encodes all these input points with a PointPillars encoder \cite{lang2019pointpillars} followed by a 2D convolutional backbone \cite{tan2020efficientnet}.
A final per-agent embedding is extracted by cropping a patch out of the feature map at a location that maps to the current location of the agent in the scene, as is in~\cite{chai2019multipath}.
Note that even though we end up with a per-agent embedding, all of the upstream processing is done for the full scene at once. The final per-agent embedding is transformed into a GMM (eqn.\ref{eqn:GMM}) using a MLP based decoder, as for the ACM.
 
\begin{figure}[h]
\centering
\includegraphics[width=0.7\columnwidth]{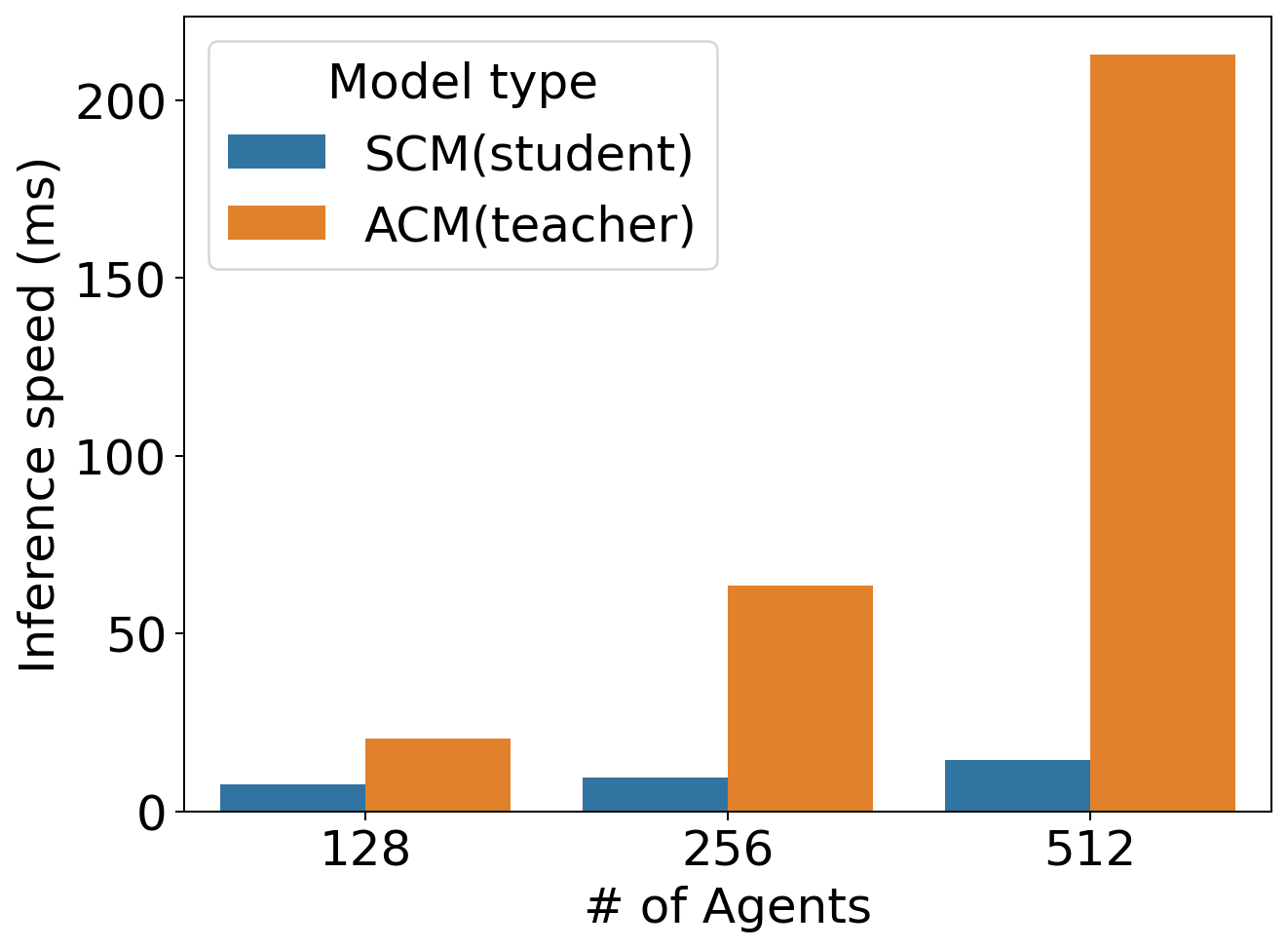}
\caption{Inference speed comparison between ACM (teacher) and SCM (student). \vspace{-15pt}}
\label{fig:inference}
\end{figure}

Fig.\ref{fig:inference} provides inference speed comparisons between SCM and ACM. As is shown in the figure, the inference speed difference gets progressively larger as the number of agents in the scene increases, showing that the ACM doesn't scale well.

Despite SCMs' fast inference speed, we observe that they underperform ACMs in general. We see this trend in public leaderboards, where agent-centric models tend to dominate (see sec.~\ref{sec:experiments}). We also see this directly comparing the ACM and SCM architectures described in this section. To get the best of both worlds (fast inference speed + good prediction accuracy), we now discuss using knowledge distillation from a slower but accurate teacher (ACM) to improve a faster but less accurate student (SCM).





\textbf{Learning Objective.}
Let the training data be in the form of $\{x^m, \hat{s}^m \}^M_{m=1}$ with $\hat{s}^m$ be the groundtruth trajectory, $\pi(s^k|x), \mu^k_t(x), \Sigma^k_t(x)$ be the outputs of a deep neural network parameterized by $\theta$.
For both the ACM and SCM, we train to maximize the log-likelihood of recorded driving trajectories, following \cite{chai2019multipath}:
\vspace{-0.3cm}\par\nobreak
{\scriptsize
  \setlength{\abovedisplayskip}{2pt}
  \setlength{\belowdisplayskip}{\abovedisplayskip}
  \setlength{\abovedisplayshortskip}{0pt}
  \setlength{\belowdisplayshortskip}{2pt}
\begin{align}
    \mathcal{L}_{base} (\theta) =& - \sum^{M}_{m=1}\sum^{K}_{k=1} \mathds{1}(k=\hat{k}^m)& \nonumber \\ & \Bigg(\log\Big(\pi(s^k|x^m;\theta)\Big)   + \sum^T_{t=1} \log\Big(\mathcal{N}(s^k_t | \mu^k_t, \Sigma^k_{t}; x^m, \theta)\Big) \Bigg), 
    \label{eqn:loss}
\end{align}
}%
where $\mathds{1}(\cdot)$ is the indicator function, $\hat{k}^m$ is the index of the predicted trajectory closest to the ground-truth $\hat{s}^m$, in terms of the L2 distance. The first term in the loss function fits the likelihood of each $k_{th}$ predicted trajectory (by making the closest-to-ground-truth predicted trajectory the most probable one), and the second term is simply a time sequence extension of standard GMM likelihood fitting \cite{bookgmm}. The advantage of training the network according to eqn.\ref{eqn:loss} is that it avoids the need of performing the  expectation-maximization procedure and avoids the intractability in directly fitting the GMM likelihood.

\section{Distillation Methods}
In this section, we describe distillation techniques we developed for trajectory-based behavior prediction\footnote[5]{Note that an alternative representation for future behavior, probabilistic occupancy grids (or ``heatmaps") could be considered. Possibly simpler distillation approaches for this representation could be developed. However heatmap representations are significantly less common in the literature, and more importantly, public benchmarks and their metrics specifically require  trajectory-based representations.}.  While we use these methods to distill from ACMs to SCMs in this work, the distillation methods here can be applied to any trajectory-based behavior prediction models. An overview of our approach is provided in Figure~\ref{fig:system_overview}.

To facilitate the presentation, we use the following notations for the teacher model. We denote $\vartheta$ as the teacher network parameters, $\xi^k$ as the $k^{th}$ predicted trajectory output from the teacher, and $\Pi(\xi^k|x)$ as the trajectory likelihood distribution from the teacher (analogous to $\pi(s^k|x)$ from the student). Lastly, we denote $\mathcal{H}(\cdot, \cdot )$ as the cross entropy function and $\mathcal{D}_{\text{KL}}(\cdot || \cdot)$ as the KL divergence function. Our distillation approaches are as follows.

\subsection{Trajectory Set Distillation}
In this distillation approach, we train our student model to match the full trajectory set output from the teacher. Recall that the full output representation of our models is a GMM; ignoring the covariances and taking the mode of each component gives us our trajectory set. The weights over this trajectory set are given by $\pi$.

The distillation loss has two parts. 
For the first part, we use the teacher's predicted trajectories (all $K$ of them) as multiple pseudo-groundtruth trajectories for training the student.
Here, we want the $k^{th}$ teacher trajectory to be maximally likely under the learned distribution for the student's corresponding $k^{th}$ mode.
For the second part, we impose a cross entropy loss to encourage the student's trajectory mode distribution $\pi$ to match the teacher's mode distribution $\Pi$.
\par\nobreak
{\scriptsize
  \setlength{\abovedisplayskip}{2pt}
  \setlength{\belowdisplayskip}{\abovedisplayskip}
  \setlength{\abovedisplayshortskip}{0pt}
  \setlength{\belowdisplayshortskip}{2pt}
\begin{align}
    \mathcal{L}_{distill}(\theta) =  - \sum^{M}_{m=1}\sum^{K}_{k=1}& \nonumber \\ \Bigg(-\mathcal{H}\Big(\pi({s}^k|x^m;\theta),    \Pi(\xi^k|x^m;&\vartheta)\Big) + \sum^T_{t=1} \log\Big(\mathcal{N}(\xi^k_t |  \mu^k_t, \Sigma^k_{t}; x^m, \theta)\Big) \Bigg)
\end{align}
}%
The full loss function is formed by adding $\mathcal{L}_{distill}$ on top of the original loss $\mathcal{L}_{base}$ (eqn.\ref{eqn:loss}) as follows:
\begin{equation}
     \mathcal{L}(\theta) = \mathcal{L}_{distill}(\theta) + \lambda \mathcal{L}_{base}(\theta) .
     \label{eqn:full_distil}
\end{equation}
Note that $\mathcal{L}_{distill}$ does not have the term $\mathds{1}(k=\hat{k}^m)$, compared to $\mathcal{L}_{base}$. This is because for the $\mathcal{L}_{distill}$, we match all $K$ predicted trajectories to the teacher's predicted trajectories, while for the $\mathcal{L}_{base}$, we only optimize over one trajectory (the real observed future groundtruth).
One added benefit of this distillation formulation is that training includes additional information in the form of additional soft labels to learn from, that is, an additional $K-1$ predicted trajectories with the associated distribution over them.

One implementation detail to note for this approach is that it imposes correspondence between each of the $K$ teacher trajectories and $K$ student trajectories, which constrains the set of possible solutions for the student by removing equivalent solutions under permutation.

We use the hyper-parameter $\lambda$ to optionally disable the base loss for a certain number of steps of warm-up, which pre-trains the model with the distillation loss only. In our experiments we cross-validate whether we (i) set $\lambda = 1$ for all training (i.e., no pre-training), or (ii) set $\lambda = \mathds{1}(\text{step} \geq \text{total steps} / 4)$, i.e., pre-train for 25\% of the total training iterations.

\subsection{Trajectory Sample Distillation}
As an alternative to using multiple trajectories as pseudo-groundtruth, as described above, we sample a single trajectory from the teacher's distribution to be the groundtruth for the student:
$\xi^k_{\text{sampled}} \sim \Pi(\xi^k |x^m, \vartheta)$. We call this sampled teacher trajectory the \emph{proxy groundtruth label}.

Then, we directly optimize over this proxy groundtruth (instead of the true groundtruth label). Mathematically, this is expressed as follows:
\par\nobreak
{\scriptsize
  \setlength{\abovedisplayskip}{2pt}
  \setlength{\belowdisplayskip}{\abovedisplayskip}
  \setlength{\abovedisplayshortskip}{0pt}
  \setlength{\belowdisplayshortskip}{2pt}
\begin{align}
    \mathcal{L}_{\text{sampled}} (\theta) = - \sum^{M}_{m=1}\sum^{K}_{k=1}&
    \mathds{1}(k=\hat{k}^m_{\text{sampled}})
     \nonumber \\ \Bigg(\log\Big(\pi(s^k|x^m;\theta)\Big) & + \sum^T_{t=1} \log\Big(\mathcal{N}({s}^k_{t} |  \mu^k_t, \Sigma^k_{t}; x^m, \theta)\Big) \Bigg)
\end{align}
}%
On expectation, over infinite samples, this loss is equivalent to requiring the student's weighted trajectory set to match the teacher's. While this formulation stands out its simplicity, it is the same as $\mathcal{L}_{base}(\theta)$, it does not, however, encourage the full GMM distribution of the teacher and student to match; this is described next.

\subsection{Trajectory Distribution Distillation}
In this last formulation, the loss directly encourages the student's full GMM output to match the teacher's GMM. As in {\em Trajectory Set Distillation}, we force correspondence between the teacher and students $k^{th}$ trajectory (for all $k$) to avoid permutation ambiguity in the solution space of the student. To match distributions, we use cross-entropy loss between the discrete mode distributions of the student and teacher ($\pi$ and $\Pi$), and KL-divergence for each Gaussian distribution ($\mathcal{N}_t$ for the student, $\mathds{N}_t$ for the teacher) in the trajectory sequences:
\par\nobreak
{\scriptsize
  \setlength{\abovedisplayskip}{2pt}
  \setlength{\belowdisplayskip}{\abovedisplayskip}
  \setlength{\abovedisplayshortskip}{0pt}
  \setlength{\belowdisplayshortskip}{2pt}
\begin{align}
    \mathcal{L}(\theta) &= \mathcal{L}_{base}(\theta)  + \sum^{M}_{m=1}\sum^{K}_{k=1} \Bigg(\mathcal{H}\Big(\pi, \Pi \Big) + \sum^T_{t=1} \mathcal{D}_{\text{KL}}(\mathcal{N}_t || \mathds{N}_t) \Bigg)
\end{align}
}%
\begin{table*}[h]
\centering
\begin{tabular}{lccccccc}
       & coord.     & test set      &                   &                      &                      &                          & \\
method & frame      & rank          & mAP($\uparrow$)   & minADE($\downarrow$) & minFDE($\downarrow$) & Miss Rate($\downarrow$)  & \\
\toprule 
Scene-Transformer\cite{ngiam2021scene}  & scene & $4^{th}$ & 0.337 & 0.678 & 1.376 & 0.198 &  \\
\best{Multipath++}                      & agent & $1^{st}$ & \best{0.401} & \best{0.569} & \best{1.194} & \best{0.143}  &  \\
\midrule                                          
ACM (teacher)                           & agent & -- & 0.329 & 0.676 & 1.488 & 0.178       &  Average \\
SCM baseline                            & scene & -- & 0.322 & 0.757 & 1.691 & 0.205        & improvement: \\
\midrule                                          
\first{SCM+Distill Set}                 & scene & $3^{rd}$ & \first{0.349} & \first{0.710} & \first{1.569} & \first{0.186}        & \first{7.8\%} \\
\second{SCM+Distill Sample}             & scene & -- & 0.320 & \second{0.742} & \second{1.643} & \second{0.194} & \second{4.2\%} \\
SCM+Distill Distr.                      & scene & -- & \second{0.330} & 0.758 & 1.681 & 0.199      & 1.5\% \\
\end{tabular}
\caption{Model distillation performance on the WOMD test set. \vspace{-15pt}}
\label{table:WOMD}
\end{table*}




\section{Experiments}
\label{sec:experiments}
We ran our experiments on the WOMD, Argoverse, and In-House datasets.
The results are shown in Table \ref{table:WOMD}, \ref{table:argoverse}, and \ref{table:in-house}. 
The best numbers across the entire table are highlighted as bold. The best methods among the SCM methods are marked as blue, and the second best ones as orange. 
In Table \ref{table:WOMD} and \ref{table:argoverse}, the first section of rows show the results of the models top-ranked in the corresponding public leaderboards.

For both the teacher and student models, we train end-to-end using the Adam optimizer with a learning rate of $5\times10^{-4}$. We used gradient clipping to prevent gradient explosion with a threshold of 10. 
We trained all models for 1M training steps, and we submit to the leaderboard the best model based on its performance on the validation set.
After cross-validation we set $\lambda = \mathds{1}(\text{step} \geq \text{total steps} / 4)$ for the WOMD and $\lambda=1$ for the In-house and Argoverse dataset. 
The teacher implementation uses off the shelf components such as polylines and LSTMs, as described in sec.~\ref{sec:acm}.
The student models use an EfficientDet-d2 backbone, a 200 x 200 PointPillars grid with a 2 meters resolution, and a PointPillars embedding size of 64.

\mypar{Metrics.} 
We follow the Argoverse benchmark and use the following metrics for evaluation: minimum average displacement error (minADE), minimum final displacement error (minFDE), and miss rate (MR). Besides these, there are additional metrics provided for each dataset: mAP (mean Average Precision) for WOMD, brier-minFDE (which scales the minFDE with prediction probability) for Argoverse, wADE (probability weighted Averaged Displacement Error) for the In-house dataset. Where the choice of $K$ is required, to define the top $K$ trajectories to be used for evaluation of a metric (for example minADE ($K=k$) on Argoverse), we use $k=6$.


\mypar{Datasets.}
{\em The Waymo Open Motion Dataset} (WOMD)~\cite{womd} is an open source data set for behavior prediction from Waymo. It contains 570 hours of data over 1750km of driving distance with more than 100,000 scenes that are on average about 20 seconds long. Our results on this dataset are summarized in Table~\ref{table:WOMD}, and we report the rank in terms of mAP.

The {\em Argoverse Motion Forecasting Competition} is a open source trajectory prediction dataset with more than 300,000 curated scenarios \cite{argoverse} with each sequence containing one target vehicle for prediction. Our results on this dataset are summarized in Table.\ref{table:argoverse}, and ranks are reported in terms of minADE.

The {\em In-House Dataset} is a large scale real-world dataset of driving scenes in various urban and suburban environments within the US. It is collected by vehicles equipped with an industry grade sensor and perception stack, and it provides detailed logs of tracked objects. Results on this dataset are a valuable addition to the public benchmark results due to its large size and quality---over 13 millions training samples with richer HD map and state information. Our distillation results on this dataset are summarized in Table.\ref{table:in-house}.

\begin{table*}[!htb]
\centering
\begin{tabular}{lccccccc}
       & coord.     & test set      &                   &                      &                      &                          & \\
method & frame & rank & brier-minFDE($\downarrow$) & minFDE($\downarrow$) & MR($\downarrow$)   & minADE($\downarrow$)   \\ \toprule 

LaneRCNN\cite{zeng2021lanercnn}         & scene & $58^{th}$ & 2.147 & 1.453 & \best{0.123} & 0.904  & \\
LGN\cite{liang2020learning}             & scene & $35^{th}$ &2.059 & 1.364 & 0.163 & 0.868 & \\
mmTransformer \cite{liu2021multimodal}  & agent  & $15^{th}$ & 2.033 & 1.338 & 0.154 & 0.844 & \\
TPCN \cite{ye2021tpcn}                  & agent & $6^{th}$ & 1.929 & {1.244} & 0.133 & {0.815} & \\
\textbf{poly}                                    & agent & $1^{st}$ & \best{1.793} & \best{1.214} & 0.132 & \best{0.790} & \\
\midrule
ACM (teacher)                           & agent & -- & {1.906}  & 1.280      & 0.147  & 0.816  & Average  \\
SCM baseline                            & scene & -- & \third{2.206} & \third{1.588}      & \third{0.225}     & \third{0.931} & improvement:      \\
\midrule
\second{SCM+Distill Set}                & scene & -- & \second{2.052} & \second{1.416}       & \second{0.180}    & \second{0.868}  & \second{11.1\%} \\
\first{SCM+Distill Sample}              & scene & $17^{th}$ &\first{2.017} & \first{1.383}    & \first{0.173} & \first{0.853}  & \first{13.2\%} \\
SCM+Distill Distr.                      & scene & -- & 2.345 & 1.723 & 0.254   & 0.980  &  -8.2\% \\

\end{tabular}
\caption{Model distillation performance on the Argoverse test set.}
\label{table:argoverse}
\end{table*}
\begin{table*}[!htb]
\centering
\begin{tabular}{lccccccc}
      & coord.           &                   &                      &                      &                          & \\
method  & frame      & wADE($\downarrow$)  & minADE($\downarrow$) & minFDE($\downarrow$) & Miss Rate($\downarrow$) \\
\toprule    
\best{ACM (teacher)}              & agent &  1.200   & \best{0.524}   & \best{1.145}   & 0.335    & Average: \\
SCM baseline                      & scene &  1.270  & \second{0.558}   & \second{1.532}   & 0.357   & improvement: \\
\midrule
\first{SCM+Distill Set}               & scene & \first{1.190}   & \first{0.545}   & \first{1.526}  &  \first{0.324}   & \first{4.6\%} \\
SCM+Distill Sample           & scene &  1.270 & 0.567   & 1.602   & \second{0.345}  & -0.7\% \\
SCM+Distill Distr.                & scene &  \second{1.220}   & 0.608   & 1.789   & 0.358   & -5.5\% \\
\end{tabular}%
\caption{Model distillation  performance on the In-House test set.}
\label{table:in-house}
\end{table*}
\begin{figure*}[!htb]
\centering
\includegraphics[width=0.89\textwidth]{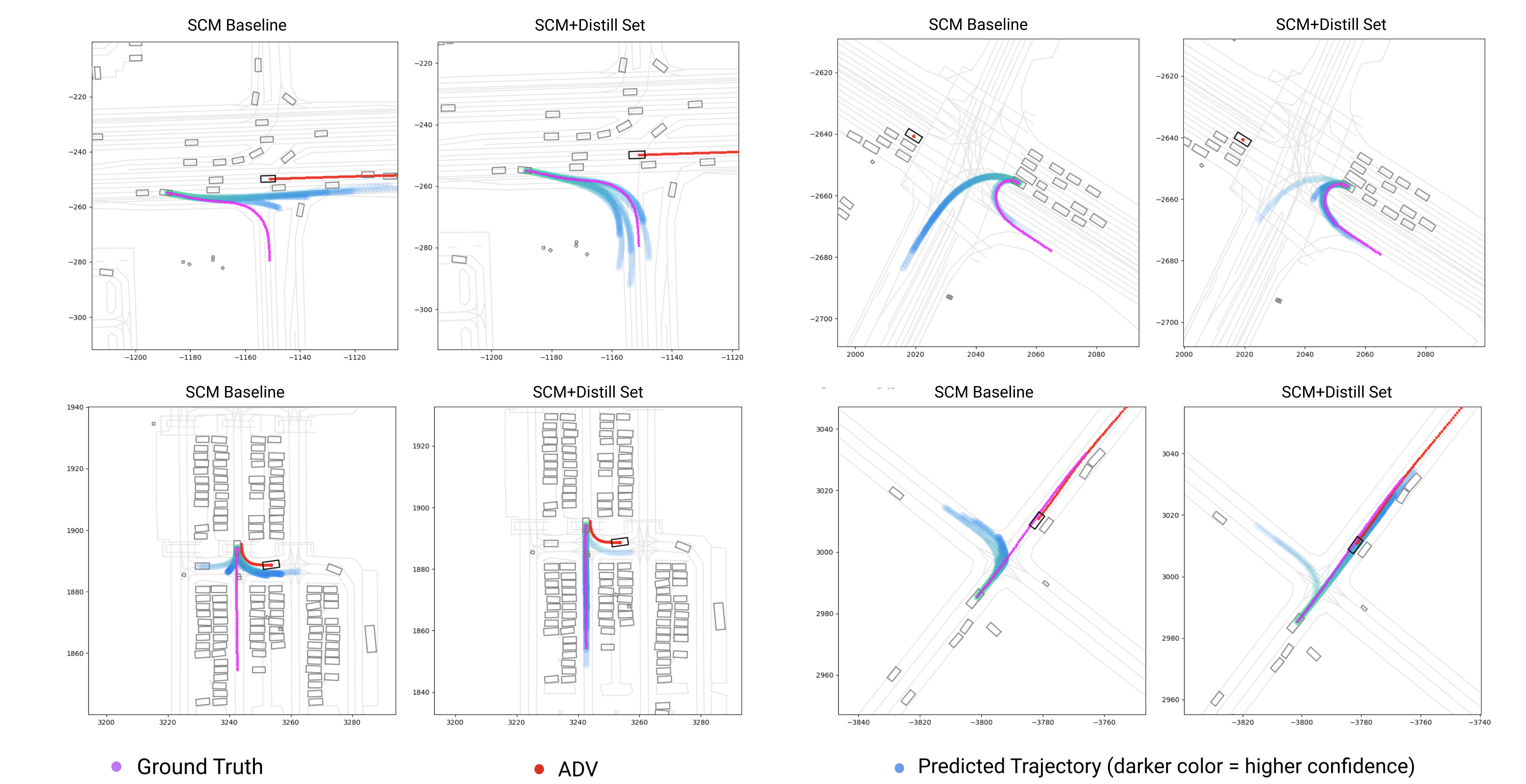}
\caption{Illustration of improvements provided by the distilled models on the WOMD. For each example, the sub-figure on the left shows prediction for SCM-baseline while the sub-figure on the right shows the predictions for our SCM+Distill Set. The purple markers show the groundtruth while the red markers show the trajectory of the Autonomous Driving Vehicle (ADV). The predicted trajectories are shown in blue (the darker the blue, the higher the confidence). In different scenarios (parking lots, variuous types of traffic intersections) we see that the non-distilled baseline misses the groundtruth trajectory and predicts left or right turn while the groundtruth is going straight, or vice versa. In contrast, after applying the distillation techniques proposed in this paper, the model more accurately predicts the groundtruth.}
\label{fig:merge2}
\end{figure*}


\subsection{Discussion}
Some clear trends emerge from our results. 
Across datasets, distillation improves the student model's performance significantly---from 4.6\%--13.2\% average relative improvement across metrics. The {\em Trajectory Set} distillation method worked the best across datasets.  Interestingly {\em Trajectory Sample} distillation worked better only on Argoverse. The major differences with Trajectory Sample distillation is that it trains with a single trajectory groundtruth rather than trying to learn a trajectory set or full distribution like the other methods. The Argoverse dataset as well stands out from other datasets in that it is smaller, has a short prediction horizon, and has less diverse driving behavior~\cite{womd}. Lastly, {\em Distribution Distillation} did not work as well as the other distillation methods across all 3 datasets.  We hypothesize that this form of distillation task was too constrained: matching GMM distributions via KL-divergence is more difficult to achieve than simply maximizing likelihood of pseudo-groundtruth (as in Trajectory Set and Trajectory Sample distillation).

Another trend from our experiments is that agent-centric models outperform the scene-centric models, in our own implementations, as well as in related works. This was our original motivation for this work, and the results presented here provide further justification for investigating distillation approaches. However, there is still improvements to be made in efficient, scene-centric models, since the gap has not been fully closed by our distillation techniques.

Lastly, we want to highlight that our models are competitive on public leaderboards in an absolute sense. On WOMD, our best distilled model\footnote[6]{Our best distillation model is named as MPG-Distil(pretrain) on the WOMD's public leader-board} is ranked $3^{rd}$, and to our knowledge is the best performing {\em scene-centric} (and thus efficient) model. In the Argoverse leaderboard, our best distilled model ranks $17^{th}$ where other known, popular {\em scene-centric } models are ranked $35^{th}$ and $58^{th}$ place.

Illustrations of improvements provided by the distilled models on a variety of urban driving scenarios are shown in Figure~\ref{fig:merge2}.
\vspace{-0.2cm}\section{CONCLUSIONS}
In this paper, we develop novel knowledge distillation techniques to bridge the coordinate-frame gap in behavior prediction models. We use a agent-centric model as a teacher to improve the accuracy of an otherwise more efficient scene-centric model. Our method improves the performance of the scene-centric model by 13.2\% on the public Argoverse benchmark, 7.8\% on the public Waymo Open Dataset, and up to 9.4\% on a large In-House dataset. The resulting improved scene-centric models are also $15$ times faster than their agent-centric distillation counterparts in busy urban scenes.

\bibliography{references}
\bibliographystyle{IEEEtran}

\end{document}